\documentclass[runningheads]{llncs}
\usepackage{graphicx}
\usepackage{dsfont}
\usepackage{amsmath}
\usepackage{float}
\usepackage{mathtools}
\usepackage[T1]{fontenc}
\usepackage{amsmath}
\usepackage{caption}
\usepackage{subcaption}
\usepackage{textalpha}
\usepackage{dsfont}
\usepackage{amsmath}
\usepackage{float}
\usepackage{multirow}
\usepackage{hhline}
\usepackage{longtable}
\usepackage{caption}
\usepackage{xcolor,pifont}
\usepackage{multicol}
\usepackage[left=17.5mm,right=17.5mm,top=24.5mm,bottom=33.95mm]{geometry}
\usepackage{tabularx}
\usepackage{amssymb}
\usepackage{pifont}
\usepackage{adjustbox}
\usepackage{lipsum}
\usepackage{tikz}
\usepackage{enumitem}
\usepackage{booktabs}
\usepackage{xcolor,pifont}
\usepackage{amssymb}
\usepackage{pifont}
\newcommand*\colourcheck[1]{%
  \expandafter\newcommand\csname #1check\endcsname{\textcolor{#1}{\ding{52}}}%
}
\usepackage{xcolor,pifont}
\colourcheck{blue}
\colourcheck{green}
\colourcheck{red}

\begin{document}

\title{Towards End-to-End Semi-Supervised Table Detection with Deformable Transformer}
\author{Tahira Shehzadi*\inst{1,2,4}\orcidID{0000-0002-7052-979X} \and
Khurram Azeem Hashmi\inst{1,2,4}\orcidID{0000-0003-0456-6493} \and
Didier Stricker\inst{1,2,4} \and
Marcus Liwicki\inst{3} \and
Muhammad Zeshan Afzal\inst{1,2,4}\orcidID{0000-0002-0536-6867}}
\authorrunning{T. Shehzadi et al.}
%
\institute{Department of Computer Science, Technical University of Kaiserslautern, 67663 Kaiserslautern, Germany \and
Mindgarage, Technical University of Kaiserslautern, 67663 Kaiserslautern, Germany \and
Department of Computer Science, Luleå University of Technology, 971 87 Luleå, Sweden \and
 German Research Institute for Artificial Intelligence (DFKI), 67663 Kaiserslautern, Germany\\
\email{\{tahira.shehzadi@dfki.de\}}}
\maketitle              

\begin{abstract}
Table detection is the task of classifying and localizing table objects within document images. With the recent development in deep learning methods, we observe remarkable success in table detection. However, a significant amount of labeled data is required to train these models effectively. Many semi-supervised approaches are introduced to mitigate the need for a substantial amount of label data. These approaches use CNN-based detectors that rely on anchor proposals and post-processing stages such as NMS. To tackle these limitations, this paper presents a novel end-to-end semi-supervised table detection method that employs the deformable transformer for detecting table objects. We evaluate our semi-supervised method on PubLayNet, DocBank, ICADR-19 and TableBank datasets, and it achieves superior performance compared to previous methods. It outperforms the fully supervised method (Deformable transformer) by +3.4 points on 10\% labels of TableBank-both dataset and the previous CNN-based semi-supervised approach (Soft Teacher) by +1.8 points on 10\% labels of PubLayNet dataset. We hope this work opens new possibilities towards semi-supervised and unsupervised table detection methods.

\end{abstract}
\section{Introduction}
A visual summary is the main aspect of different applications in document analysis, such as recognizing graphical components in the visualization pipeline and summarizing the content of a document. As a result, localizing and detecting graphical items such as tables will be an important action in the analysis and summary of the document. Due to the increase in the number of documents, manually retrieving the table data is no longer practical. Automated processes offer efficient, reliable, and successful solutions for manual tasks. Previously, optical character recognition \cite{ocr8,ocr4} and rule-based \cite{RecogTable5,TTsurvey8,extractTab9} table detection approaches  were used to identify and locate them. Then, some automated methods \cite{tab6f5,tabgd4,khurb5} have been suggested to detect tables. However, these approaches are often rule-based because the documents have a set structure or dimension \cite{GOjt3}. Moreover, they cannot generalize to a new table structure, such as borderless tables. Later on, deep learning methods were used by researchers to identify them \cite{fast15,faster23,YOLO9000,mask-rcnn84}, and shows that machine-learning approaches are more effective than traditional methods \cite{mlr7}. 

 Deep learning approaches \cite{DeepDeSRT3,TDcl34,TSRkhurm4,cas10,rethink78,naik86} do not rely on rules and can accurately generalize the problem. However, deep learning models take a considerable quantity of labeled data for training. These supervised methods achieve impressive results on public benchmarks, and their performance cannot be translated into industrial applications unless similar large-scale annotated datasets exist in that domain. It is potentially error-prone and time-consuming to generate this data manually or via other pre-processing approaches. Therefore, it is important to develop a semi-supervised approach due to concerns about the availability of labeled training data, which shifts the problem from a supervised to a semi-supervised setting. Recently, semi-supervised learning-based methods are introduced in computer vision containing two detectors. The first detector provides pseudo labels for unlabeled data. The second detector trains using pseudo labels generated by the first detector and a small percentage of label data and provides final predictions. Both detectors update each other during training. This approach has been described in several works, including \cite{selfsup87,propsemi6,activsemi3,unsupaug4}. In most cases, the first detector is not strong enough, which can negatively impact the pseudo-labeling process. Moreover, previous semi-supervised approaches used CNN-based networks \cite{faster23} that depend on anchors to generate region proposals and post-processing stages such as Non-Maximal suppression (NMS) to reduce the number of overlapping predictions. 
 
To address these limitations, this paper proposes a semi-supervised table detection approach that employs the deformable transformer \cite{Deformable54}. It generates pseudo-labels for unlabeled data and then trains the detector using them and a small quantity of label data in each iteration. This approach aims to improve the pseudo-label generation procedure by iteratively refining the pseudo-labels and the detector. It involves training in two modules. The teacher module contains a pseudo-labeling framework. The student module is the final detection network that uses pseudo-labels and a small quantity of label data. The teacher module is simply an Exponential Moving-Average (EMA) of the student module, which ensures that the pseudo-label generation and detection modules are constantly updating each other. Unlike other pseudo-labeling methods, we propose the idea of employing the deformable transformer that allows completing the pseudo-labeling process without needing object proposals and post-processing steps as Non-maximal suppression (NMS).  Another benefit is having a dynamic effective receptive field to adapt fot tables of different sizes and scales in the input image. This allows the network to effectively detect tables of varying sizes and orientations, making it more robust and versatile. Additionally, this framework has a reinforcing effect, providing that the Teacher model consistently monitors the Student model. In this paper, we show through empirical evidence that this semi-supervised table detection approach that uses a deformable transformer can produce results comparable to CNN-based approaches without needing object proposals and post-processing steps such as Non-maximal suppression (NMS). 

\noindent In summary, the main contributions of the paper are as follows:
\begin{itemize}
\item[$\bullet$] We present an end-to-end semi-supervised table detection method that employs the deformable transformer and allows completing the pseudo-labeling process without needing object proposals and post-processing steps such as Non-maximal suppression (NMS).
\item[$\bullet$]  We formulate the problem of table detection as an object detection problem and leverage the potential of deformable detection transformer for this task. To the best of our knowledge, this work is the first that exploits the transformer-based method in a semi-supervised setting.
\item[$\bullet$] We perform an exhaustive evaluation on four different datasets, PubLayNet, DocBank, ICDAR-19 and TableBank, and produce results comparable to CNN-based semi-supervised approaches without needing object proposals process and post-processing steps such as NMS.
\end{itemize}
\section{Related Work}
\label{sec:Literature-Review}
Table detection is an essential task for document image analysis. Many researchers have proposed different approaches for detecting tables containing arbitrary structures in document images. Previously, most presented approaches used custom rules or relied on extra meta-data input to deal with table detection tasks \cite{tsk5,tupextract3,strctRTB3,DPmatch4}. Recently, researchers employed statistical methods \cite{tsr87} and deep learning approaches to make the table detection systems more generalizable \cite{DeepDeSRT3,DeCNT82,CasTab45,Hyb65}. This section gives a detailed summary of these techniques and an overview of the CNN-based semi-supervised object detection methods.
\subsection{Rule-based Approaches}
To the best of our knowledge, Itonori et al. \cite{tsk5} presented a table detection approach for the first time on document images. This method represents the table as a text block that uses specified rules. Later, \cite{strctRTB3} introduced a table detection approach that works on horizontal and vertical lines. Pyreddy et al. \cite{TINTIN67} proposed a procedure that extracts tabular regions from the text using custom heuristics. Pivk et al. \cite{PIV67} presented a system that transforms HTML format table documents into logical forms. It introduces an appropriate tabular layout employed for extracting tables. Hu et al. \cite{DRR6}  presented a table detection approach that relies on white regions and vertically connected elements in document images. Readers can find a complete overview of these rule-based methods in \cite{RecogTable5,TTsurvey8,extractTab9,DEA38,TSsurvey32}. Though rule-based approaches perform fine on document images with matching table formats, these methods can not provide generic solutions. Therefore, systems with more generalizable abilities are needed to solve table detection tasks on document data.

\subsection{Learning-based Approaches}
Cesarini et al. \cite{trainTD5} presented a supervised learning system for detecting table objects in document images. It converts a document image into an MXY tree model and labels the blocks as tables confined in horizontal and vertical lines. Hidden Markov Models \cite{RichMM3,FTabA4} and the SVM classifier with traditional heuristics \cite{lineln6} are applied to document images for table detection. Though these machine learning approaches performed better than ruled-based approaches on documents, these methods need additional information, such as ruling lines. Deep Learning-based approaches outperformed traditional approaches in accuracy and efficiency. These methods are categorised into object detection, semantic segmentation, and bottom-up approaches.

\noindent\textbf{Semantic segmentation-based Approaches.} These approaches \cite{Xi17,He761,Ik36,Paliw9} consider table detection a segmentation task and apply available semantic segmentation networks to generate segmentation masks on the pixel level and then combine table regions to provide final table detection. These methods performed better than traditional approaches on several benchmark datasets \cite{icdar19,PubLayNet3,iiit13k,icdar13,icdar17,tablebank8,pubtables5}. Yang et al. \cite{Xi17} presented a fully convolutional network (FCN) \cite{FCNseg4} for page object segmentation, which combines linguistic and visual features to enhance segmentation results for table and other page object detection. He et al. \cite{He761} presented a multi-scale FCN that provides segmentation masks table/text areas and their related contours and then refined to get final table blocks. 

\noindent\textbf{Bottom-up Approaches}
These approaches consider table detection as a graph-labeling task and define graph nodes as page objects and graph edges connection between page objects. Li et al. \cite{Li64} extracted line areas using the classic layout analysis approach, then used two CNN-CRF networks to categorise them into four categories: text, figure, formula and table and then provided a prediction of the corresponding cluster for pair of line areas. Holecek et al. \cite{Martin66} and Riba et al. \cite{busmessg4} considered text areas as nodes, formed a graph to determine the design per document and then employed graph-neural networks for node-edge classification. These approaches rely on specific assumptions, such as text line boxes as an extra input. 

\noindent\textbf{Object Detection-based Approaches.}
Detecting tables from document images can be represented as an object detection task, with table objects treated as natural objects. Hao et al. \cite{Hao789} and Yi et al. \cite{Yi77}  applied R-CNN for detecting tables, but the performance of these approaches still relies on heuristic rules as in previous methods. Later, more efficient single-stage object detectors like RetinaNet \cite{retinaNet68} and YOLO \cite{yolos6} and two-stage object detectors like Fast R-CNN \cite{fast15}, Faster R-CNN \cite{faster23}, Mask R-CNN \cite{mask86}, and Cascade Mask R-CNN \cite{cascadercnn8} were applied for other document objects such as figures and formulas detection in document images \cite{Pog84,Azka62,yolotab5,gte9,Saha43,Ayan29,Agarwal52,DeepDeSRT3,TDcl34,TSRkhurm4}. Furthermore, \cite{Azka62,Ayan29,arif48} applied different image transformation approaches, such as coloration and dilation, to improve the results further. Siddiqui et al. \cite{Sidd32} combined deformable-convolution and  RoI-Pooling \cite{deformconv4} into Faster R-CNN to provide a more efficient network for geometrical modifications. Agarwal et al. \cite{Agarwal52} used a composite network \cite{CBNet5} as a backbone with deformable convolution to increase the performance of two-stage Cascade R-CNN. These CNN-based object detectors have a few heuristic stages, like proposals generating step and post-processing steps such as non-maximal suppression (NMS). Our semi-supervised approach considers detection a set prediction task, eliminating the anchor generation and post-processing stages such as NMS and providing a simpler and more efficient detection pipeline.
\subsection{Semi-supervised Object Detection}
Semi-supervised learning approaches in object detection are divided into two types: consistency-based approaches \cite{consemi6,propsemi8} and pseudo-label generation-based approaches \cite{omnisup8,rethink98,selfsup6,semimask1,Ksemi76,simplesemi76,minsemi8}. Our method falls into the pseudo-label type. Previous work \cite{omnisup8,rethink98} combined prediction results from varied data augmentation techniques to produce pseudo-labels for unlabeled data, while \cite{selfsup6} trained a SelectiveNet to generate the pseudo-labels. In \cite{selfsup6}, a box from unlabeled data was placed onto labeled data and evaluated localization consistency on the labeled images. However, this method requires a very complex detection procedure due to the modification of the image. STAC \cite{simplesemi76} presented to perform strong augmentation on the data for pseudo-label generation and weak augmentation for model training. We propose an end-to-end semi-supervised table detection method that employs the deformable transformer. Similar to other pseudo-label generation approaches \cite{omnisup8,rethink98,selfsup6,simplesemi76,minsemi8}, it follows a multi-level training mechanism. It effectively avoids the need for anchors generation stage and post-processing steps such as Non-Maximal suppression (NMS).

\section{Methodology}
\label{sec:method}
First, we revisit Deformable DETR, a modern transformer-based object detector, in Section~\ref{sec:Deform-detr}. Later, we explain the proposed semi-supervised learning mechanism and its pseudo-label generation module in Sections~\ref{sec:semi-sup}.
\subsection{Revisiting Deformable DETR}
\label{sec:Deform-detr}
\begin{figure*}
\centering
\includegraphics[width=0.8\textwidth]{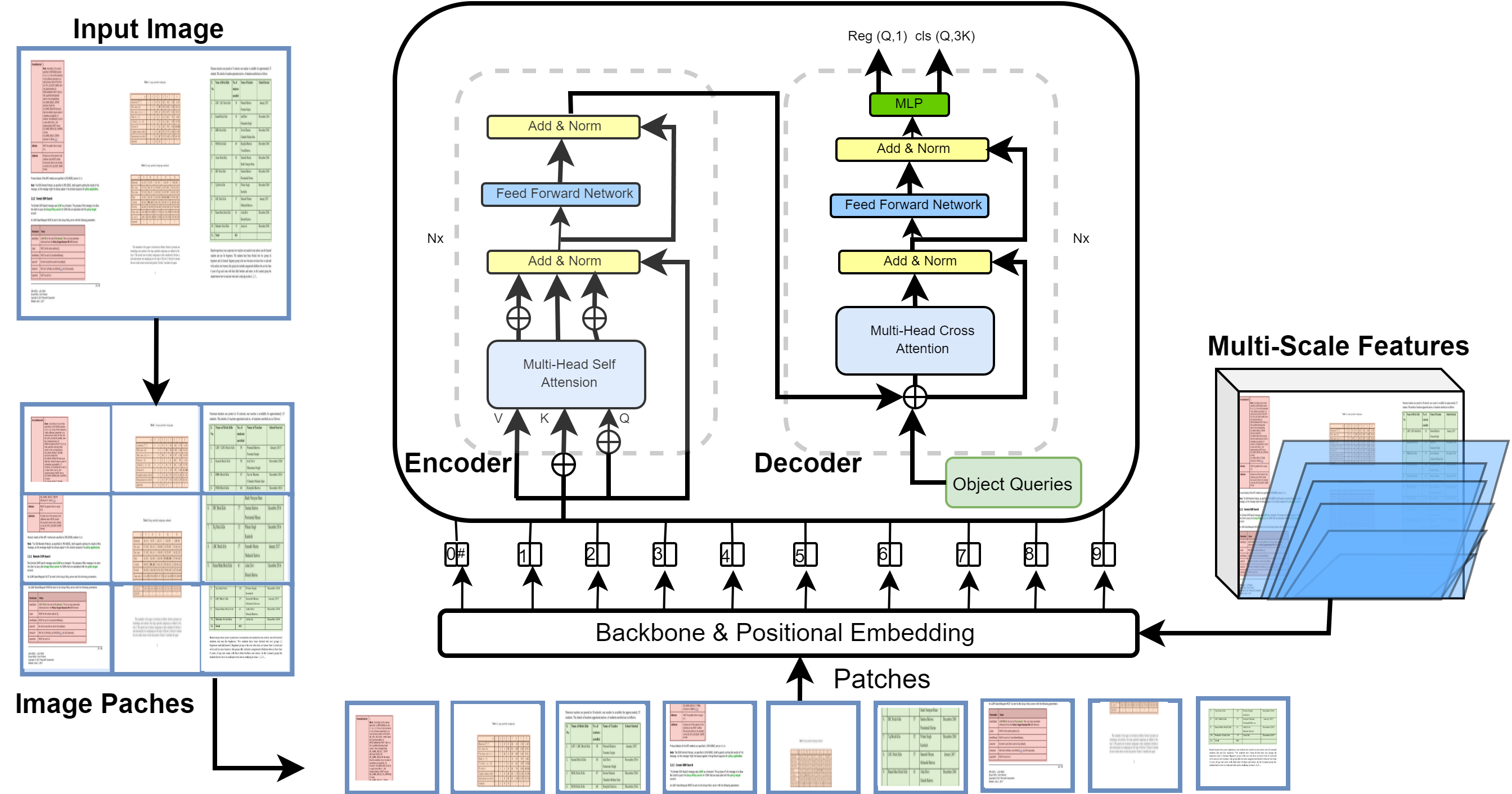}
\caption{Illustration of the deformable transformer employed in semi-supervised table detection method. We split the input image into small equal-sized patches, add position embeddings, and feed the resulting patches along with input multi-scale features to the transformer encoder. In the decoder, We use object queries as reference points and provide bounding boxes predictions and class labels as the final output.}\label{fig:encoder-decoder}
\end{figure*}

Deformable DETR \cite{Deformable54} contains a Transformer encoder-decoder network that considers object detection as a set-predictions task. It uses Hungarian loss and avoids overlapped predictions for ground-truth bounding boxes through bipartite matching. It eliminates the need for hand-crafted elements such as anchors and post-processing stages such as Non-maximal suppression (NMS) used in CNN-based object detectors. Deformable DETR is an extension of the DETR \cite{detr34}  architecture that addresses some of the limitations of DETR, such as slow training convergence and poor performance on small objects. Deformable DETR introduces deformable convolutions into the architecture, which allows for more flexible modeling of object shapes and better handling of objects of varying scales. This can lead to improved performance, particularly on small objects, and faster convergence during training. Here, we provide an overview of the encoder-decoder network, Multi-scale Feature processing and attention mechanism of deformable DETR. Figure~\ref{fig:encoder-decoder} shows all modules of the deformable transformer, including multi-scale features and encoder-decoder network.  \newline

\begin{figure}
\centering
\includegraphics[width=0.9\textwidth]{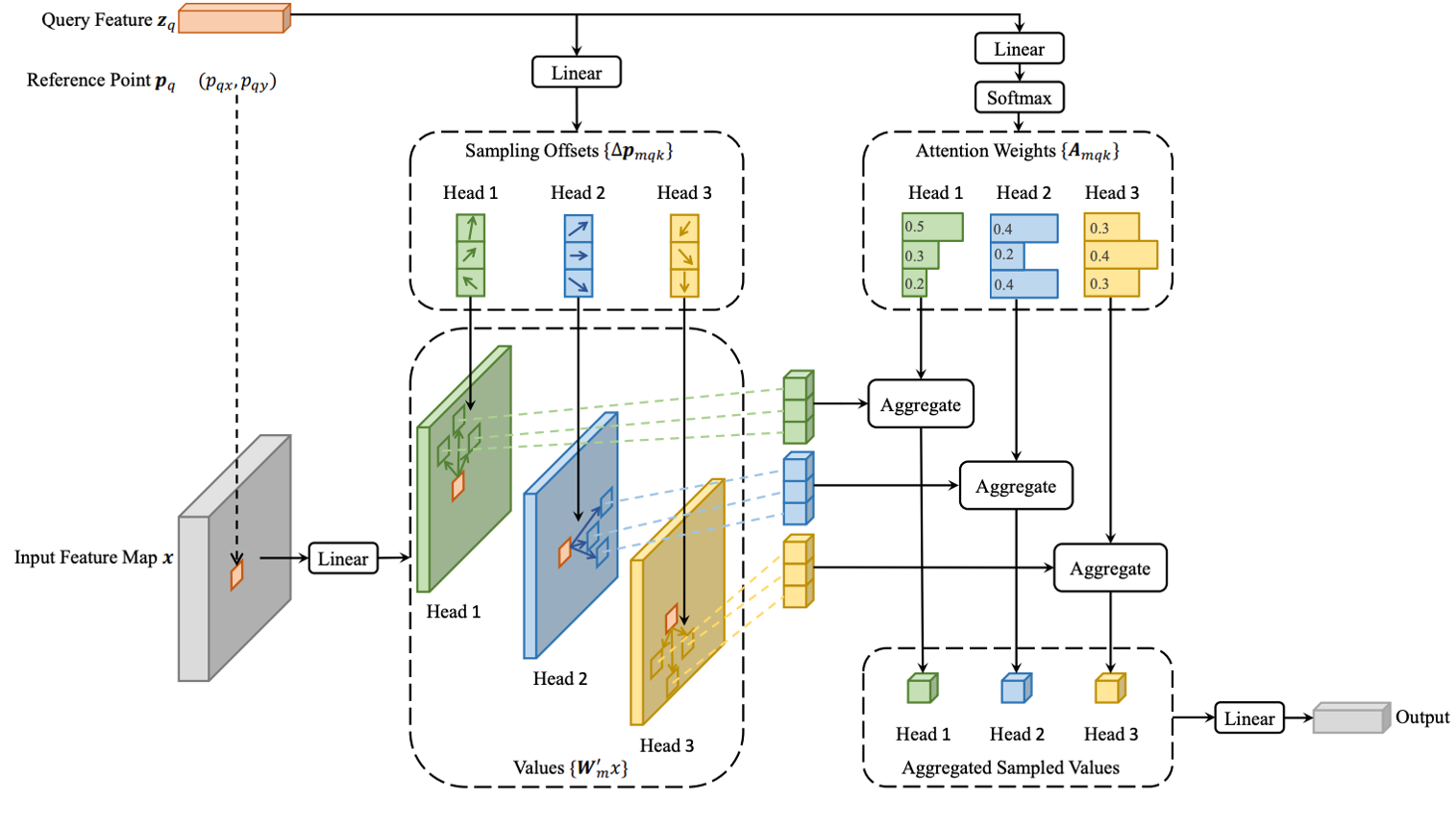}
\caption{Deformable Attention network. It considers only a few sample pixels near a reference pixel, whatever the size of input features. The query matrix takes only a small set of keys, which resolves the slow training convergence issue of DETR. (image from \cite{Deformable54}). 
}\label{fig:Deform-detr}
\end{figure}

\noindent\textbf{Transformer Encoder.} The CNN backbone (ResNet-50) extracts the input feature maps $ f_m \in \mathds{R}^{h_i \times w_i \times c_i} $. The spatial dimensional feature maps are converted into one-dimensional $ z_m \in \mathds{R}^{h_i \times w_i \times d_1} $ feature maps as the transformer encoder network takes input as a sequence. This one-dimensional vector is fed as input along with positional embeddings \cite{imgtrans45,attenaug} to the transformer encoder network, which further transforms them into features for object queries. Every layer of the encoder module contains an attention network and a feed-forward network (FFN) where query and key values are the pixels of feature maps. Readers can refer to \cite{att75} for a detailed explanation of transformer.\newline
\noindent\textbf{Transformer Decoder.}
The decoder network takes the output of the encoder features and N number of object queries as input. It contains two attention types self-attention and cross-attention. The self-attention module finds the connection between object queries. Here both key and query matrics contain object queries. The cross-attention module extracts feature using object queries from the input feature map. Here key matrix contains the feature maps provided by the encoder module, and the query matrix is the object queries fed as input to the decoder. After the attention modules, feed-forward networks (FFN) and linear projection layers are added as the prediction head. The linear projection layer predicts class labels, while FFN provides final bounding-box coordinate values.\newline
\noindent\textbf{Deformable Attention Module.}
The attention module in the DETR network considers all spatial locations of the input feature map, which makes the training convergence slower. However, a deformable DETR can solve this issue using the deformable convolution-based \cite{deformconv4,deformB3} attention network and multiscale input features \cite{fpn49,msf5}. It considers only a few sample pixels near a reference pixel, whatever the size of input features, as illustrated in Figure~\ref{fig:Deform-detr}. The query matrix takes only a small set of keys, which resolves the slow training convergence issue of DETR. Readers can refer to \cite{Deformable54} for a detailed explanation of Deformable DETR. 
\subsection{Semi-Supervised Deformable DETR }
\label{sec:semi-sup}
\begin{figure}
\centering
\includegraphics[width=0.9\textwidth]{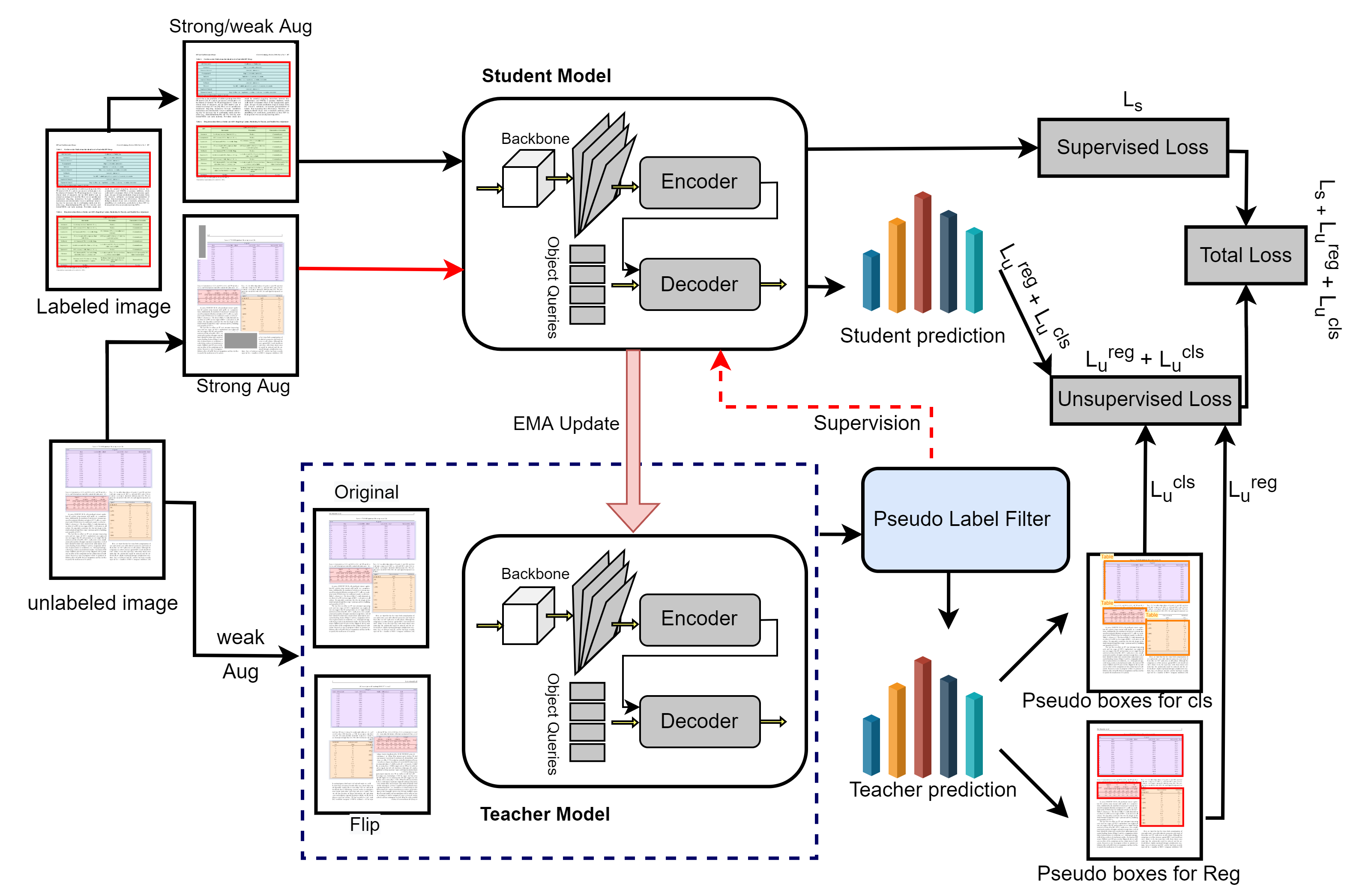}
\caption{Our proposed semi-supervised approach that employs Deformable transformer \cite{Deformable54}. (1) The training data has two data types label data and unlabeled data. (2) It contains two modules a student module and a teacher module. (3) The teacher module only takes unlabeled images as input after applying weak augmentation. (4) After applying strong augmentation on unlabeled data type, the student module takes both labeled and unlabeled images as input. (5) During training, the student module continuously updates the teacher module with an Exponential Moving-Average (EMA) \cite{sma7} strategy.}\label{fig:semi}
\end{figure}

\begin{figure*}
\centering
\includegraphics[width=0.9\textwidth]{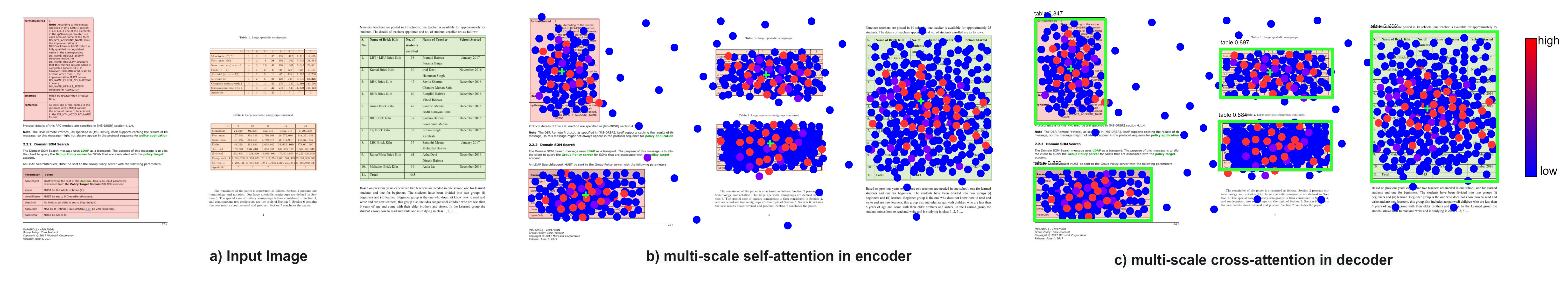}
\caption{Visualization of the sample points and attention weights from multi-scale deformable attention feature maps. Each sample point is denoted as a circle whose color represents its relative attention weight value. The reference points are the object queries taken as input in the encoder, represented by the green plus sign. In the decoder, the final bounding boxes are represented as green rectangles, and the class label and its confidence value are shown on the upper side in black text.}\label{fig:attention}
\end{figure*}

In this subsection, we describe the learning mechanism of our proposed semi-supervised approach that employs the Deformable transformer and then explain the pseudo-labeling strategy. Semi-supervised Deformable-DETR is a unified learning approach that uses fully labeled and unlabeled data for object detection. It contains two modules a student module and a teacher module. The training data has two data types label data and unlabeled data. The student module takes both labeled and unlabeled images as input where strong augmentation is applied on unlabeled data while both (strong and weak augmentation) is applied on label data. The student module is trained using detection losses of labeled and unlabeled data through pseudo-boxes. The unlabeled data contains two groups of pseudo boxes for providing class labels and their bounding boxes. The teacher module only takes unlabeled images as input after applying weak augmentation. Figure~\ref{fig:semi} presents a summary of proposed pipeline. The teacher module feeds prediction results to the pseudo-labeling framework to get pseudo-labels. Then, the student module uses these pseudo-labels for supervised training. Here, weak augmentation on unlabeled data is used for the teacher module to generate more precise pseudo-labels. Strong augmentation on unlabeled data is used for the student module to have more challenging learning. The student module also takes a small percentage of labeled images with strong and weak augmentation as input. The student module $s_m$ is optimized with the total loss as follows:
\setlength{\abovedisplayskip}{3pt}
\setlength{\belowdisplayskip}{3pt}
\begin{equation}
\mathcal{L}^{s_m} = \sum_{n} \mathcal{L} (x_j^{l,s_a}, y_j^{l,s_a})+ \mathcal{L} (x_j^{l,w_a}, y_j^{l,w_a}) + \sum_{n} \mathcal{L}(x_j^{u,s_a}, y_j^{t_m}) 
\end{equation}
Where $s_a$ represents strong augmentation, $w_a$ represents weak augmentation. $x_j^{l,s_a}$ is the strong augmented input image and its corresponding label is $y_j^{l,s_a}$. The term $x_j^{l,w_a}$ is the weak augmented input image and its corresponding label is $y_j^{l,w_a}$. For the labeled images, strong and weak augmentations are also applied for learning, and are fed to the student module. The term $x_j^{u,s_a}$ represents unlabeled strong augmented image fed to student module and the term $y_j^{t_m}$ is the pseudo-label from teacher module. Here, $\mathcal{L}$ is the weighted sum of classification (class labels) and regression (bounding box) loss as follows: 
\begin{equation}
\mathcal{L} = \alpha_1 \mathcal{L}^{reg} + \alpha_2 \mathcal{L}^{cls}
\end{equation}
Where $\alpha_1$ and $\alpha_2$ are the weight values, the teacher-student modules are initialized randomly at the start of training. During training, the student module continuously updates the teacher module with an Exponential Moving-Average (EMA) \cite{sma7} strategy.
Pseudo-label generation for image classification tasks is easy, considering probability distribution as Pseudo-labels. In contrast, object detection tasks are more complicated as an image may include numerous objects, and annotation contains object location and class label. The CNN-based object detectors use anchors as object proposals and remove redundant boxes by post-processing steps such as non-maximal suppression (NMS). In contrast, transformers use attention mechanisms and object queries. Figure~\ref{fig:attention} shows sample points and attention weights from multi-scale deformable attention feature maps for both student and teacher networks. 
Its training complexity is $O(N_q c_i^2 + min(h_iw_i c_i^2, N_q kc_i^2) + 5N_q kc_i + 3N_q c_ip_s k)$. This takes into account the computation of the sampling coordinate offsets and attention weights, as well as the bilinear interpolation and weighted sum in the attention mechanism. $N_q$ is the number of query elements, $c_i$ is the channel dimension, $k$ is the kernel size, $p_s$ is the number of sampling points, and $h_iw_i$ is the height and width of the feature map. In our experiments, $p_s = 8, k \leq 4$ and $ c_i = 256$ by default, thus $5k + 3p_s k < c_i$ and the complexity is of $ O(2N_q c_i^2 + min(h_iw_i c_i^2, N_q kc_i^2))$.
When used in the DETR encoder with $N_q = h_iw_i$, the complexity of the deformable attention module is $O(h_iw_i c_i^2)$, which scales linearly with the spatial size. When used in the DETR decoder with $N_q = N$ (the number of object queries), the complexity becomes $O(N kc_i^2)$, which is independent of the spatial size as attention is focused on the object queries.

\noindent\textbf{Training}
The semi-supervised network is trained in two steps: a) train the student module independently on labeled data and generate pseudo-labels by teacher module; b) combine training of both modules to provide final prediction results.

\noindent\textbf{Pseudo-Labeling Framework}
We used a simple framework to provide pseudo-labels for unlabeled data at the output of the teacher module, as applied in SSOD \cite{unbiasedT36}. Usually, object detectors give confidence score vector $s_k \in [0, 1]^{C_i}$ for every provided bounding box $b_k$. A simple approach to provide pseudo-labels is to just thresholding these scores. In a simple pseudo-labeling filter, pseudo-labels can be formed by providing a threshold to the confidence value $s^{c_k}_k$ of the ground-truth class $c_k$. If the prediction value is not greater than the confidence value for a ground-truth class, the highest prediction value is considered the pseudo-label.
Inspired by DETR \cite{detr34}, we develop the pseudo-label assignment task as a bipartite matching task between the teacher module predictions and the generated semi-labels. Specifically, the permutation of K elements is as follows:
\begin{equation}
 \hat{\sigma} =\arg\min_{\sigma \in N} \sum_{k}^{N_i} \mathcal{L}_{match} (y_{k},\hat{y}(k)),
\end{equation}
Where $\mathcal{L}_{match} (y_k,\hat{y}(k))$ is the match-cost between teacher labels and ground-truth semi-labels as follows:
\begin{equation}
\mathcal{L}_{match} (y_k,\hat{y}(k))=-\mathds{1}_{\{c_k\neq \phi\}}\hat{p}_{\sigma(k)}(c_k)+\mathds{1}_{\{c_k\neq \phi\}}\mathcal{L}_{bbox}(b_k,\hat{b}_{\hat{\sigma}}(k))
\end{equation}
The Pseudo-Labeling framework is applied to the predictions of teacher module $\hat{y}(k)$ where $\hat{y}(k) =\{\hat{y}^{class}, \hat{y}^{bbox}\}$ is the prediction, with $\hat{y}^{class}$ and $\hat{y}^{bbox}$ represent the class and box values, respectively. Here, $\hat{y}^{cls} = [v_1, ..., v_N ]^T \in \mathbb{R}^{N \times C_i} $ and $\hat{y}^{box} = [\hat{b_1}, ..., \hat{b_N} ]^T \in \mathbb{R}^{N \times 4}$, where $v_N$ is the output vector (before the softmax) , $\hat{b_N}$ the related bounding-box prediction, and N is the object queries provided as input to the transformer decoder. $y_{k}$ represents pseudo-labels generated from confidence-score. The optimal selection is allowed with the Hungarian match mechanism \cite{detr34,HungAP76}, giving pseudo-labels $\{(b_k , c_k )\}$. This approach to select matching between the teacher module's prediction and semi-labels generated by providing threshold works in the same way as the heuristic selection rules used for matching proposals \cite{faster23} or anchors \cite{fpn49} with ground-truth objects in CNN-based object detectors. The main difference is that it determines one-to-one matching without duplicates. The second stage calculates the loss function, the Hungarian loss for all pair matching in the last stage. We define the loss similar to the previous object detector's losses as a linear combination of a negative log-likelihood for class label and a bounding box as follows:
\begin{equation}
 \mathcal{L}_{H}(y, \hat y)= \sum_{i=1}^{N}[-log\hat{p}_{\hat{\sigma}(k)}(c_k)+\mathds{1}_{\{c_k\neq \phi\}}\mathcal{L}_{box}(b_k,\hat{b}_{\hat{\sigma}}(k))]
\end{equation}
Here, $b_k \in \mathbb{R}^4$ is the pseudo-bounding box, and $c_k$ is the pseudo-class label. $\hat{\sigma}$ is the matching determined in the previous stage. In training, we reduce the weight of log probability by ten times when $c_k$ for class imbalance. This mechanism is similar to the Faster R-CNN training strategy to balance proposals by sub-sampling \cite{faster23}.

\section{Experimental Setup}
\label{sec:exp}
\subsection{Datasets}
\label{sec:dataset}
\noindent\textbf{TableBank:} TableBank \cite{tablebank8} is the second-largest dataset in the document analysis domain for the table recognition problem. The dataset has 417,000 document images annotated through the arXiv database crawling procedure. The dataset features tables from three categories of document images: LaTeX images (253,817), Word images (163,417), and a combination of both (417,234). It also includes a dataset for recognizing the structures of the table. In our experiment, We only used the dataset for table detection from TableBank.

\noindent\textbf{PubLayNet:}
PubLayNet \cite{PubLayNet3} is a large public dataset with 335,703 images in the training set, 11,240 in the validation set, and 11,405 in the test set. It includes annotations such as polygonal segmentation and bounding boxes of figures, lists titles, tables, and text of images from research papers and articles. The dataset was evaluated using the coco analytic technique \cite{coco14}. In our experiment, we only used 102,514 of the 86,460 table annotations.

\noindent\textbf{DocBank:} DocBank \cite{Li49} is a large dataset of over 5,000 annotated document images from various sources designed to train and evaluate tasks such as text classification, entity recognition, and relation extraction. It includes annotations of title, author name, affiliation, abstract, body text, etc. 

\noindent\textbf{ICDAR-19:} The competition for Table Detection and Recognition (cTDaR) \cite{icdar19} is organized at ICDAR in 2019. For the table detection task (TRACK A), two new datasets (modern and historical) are introduced in the competition. For direct comparison against the prior state-of-the-art \cite{Ayan29}, we provide results on the modern datasets with an IoU threshold ranging from 0.5–0.9. 
\subsection{Evaluation Criteria}
\label{sec:Eval}
We use some evaluation metrics to analyze the performance of our semi-supervised table detection approach that employs the deformable transformer. This section defines the employed evaluation metrics as precision, Recall, and F1-score. The Precision \cite{pr61} is the fraction of actual instances as True Positives among the predicted instances as False Positives and  True Positives). The Recall \cite{pr61} is the fraction of actual instances as True Positives that were retrieved (True Positives + False Negatives). The F1-score \cite{pr61} is the harmonic mean of Precision and Recall. We compute the intersection over union(IoU) by performing the intersection divided by the union for the region of the ground-truth box $A_{g}$ and the formed bounding box $A_{p}$. 
\begin{equation}
IoU= \frac{area(A_{g} \cap A_{p})}{area(A_{g}\cup A_{p})} 
\end{equation}
IoU estimates that either a detected table object is a false positive or a true positive. We find the average precision(AP) by a precision-recall (PR) curve following the context of MS COCO \cite{coco14} evaluation. It is the area under the PR curve, calculated using the following equation: 
\begin{equation}
AP= \sum_{k=1}^N (Re_{k+1}-Re_k)P_{intr}(Re_{k+1})
\end{equation}
Where Re1, Re2, . . . , $Re_k$ represent the recall parameter.
The mean average precision (mAP) is often used to evaluate the performance of detection methods. It is calculated by taking the mean of average precision for all classes in a dataset. The mAP can be affected by changes in the performance of individual classes due to class mapping, which is a limitation of this metric. We set the intersection over union (IoU) threshold values at 0.5 and 0.95. The mAP is calculated as follows:  
\begin{equation}
mAP= \frac{1}{S}\sum_{s=1}^SAP_s  
\end{equation}
Where S represents total classes.
\subsection{Implementation Details}
\label{sec:implement}
We use the Deformable DETR \cite{Deformable54} with a ResNet-50 \cite{resnet45} backbone pre-trained on the ImageNet \cite{NId6} dataset as our detection framework for evaluating the usefulness of the semi-supervised approach. We perform training on PubLayNet, ICDAR-19, DocBank and all three splits of the TableBank dataset. We use 10$\%$, 30$\%$ and 50$\%$ of labeled data and the rest as unlabeled data. The threshold value for pseudo-labeling is set at 0.7. We set the training epochs to 150 for all experiments with the learning rate reduced by a factor of 0.1 at the 120th epoch. We follow \cite{unbiasedT36,Deformable54} to apply strong augmentation as horizontal flip, resize, remove patches, crop, grayscale and Gaussian blur. We use horizontal flipping to apply weak augmentation. The value N for the number of queries to the input of the decoder of Deformable DETR is set to 30 as it gives the best results. Unless otherwise specified, we evaluated the results using the mAP (AP50:95) metrics. All models are trained with a batch size of 16, using the same hyperparameters as Deformable DETR \cite{Deformable54}. The weight $\alpha_1$ is 2 and $\alpha_2$ is 5 to balance the classification loss ($L_{cls}$) and regression loss ($L_{box}$). To make the training faster, we set the height and width of the input image to 600 pixels. We employ the standard size of 800 pixels for comparison with other approaches.
\section{Results and Discussion}
\label{sec:results}
\subsection{TableBank}
In this subsection, we provide the experimental results on all splits of the TableBank dataset on different percentages of label data. We also compare the transformer-based semi-supervised approach with previous deep learning-based supervised and semi-supervised approaches. Furthermore, we give results on 10\% TableBank-both data split for all IoU threshold values. Table~\ref{tab:tablebank} provides the results of semi-supervised approach that employs deformable transformer for TableBank-latex, TableBank-word, and TableBank-both data splits on 10\%, 30\% and 50\% label data and the rest as unlabeled data. It shows that the TableBank-both data split has the highest $AP_{50}$ value of 95.8\%, TableBank-word has 93.5\%, and TableBank-both has 92.5\% at 10\% label data.

\begin{table}
\begin{center}
\caption{Performance of the semi-supervised approach that employs deformable transformer for TableBank-latex, TableBank-word, and TableBank-both data splits on different percentages of label data. Here, mAP represents mean AP at the IoU threshold range of (50:95), $AP_{50}$ indicates AP at the IoU threshold of 0.5, and $AP_{75}$ denotes AP at the IoU threshold of 0.75. $AR_L$ indicates average recall for large objects.}\label{tab:tablebank}
\renewcommand{\arraystretch}{1} 
\begin{tabular*}{\textwidth}
{@{\extracolsep{\fill}}llllll@{\extracolsep{\fill}}}
\toprule
\textbf{Dataset} &
\textbf{Label-percent} &
\textbf{mAP} & 
\textbf{AP\textsuperscript{50}} &
\textbf{AP\textsuperscript{75}}  & 
\textbf{AR\textsubscript{L}}  \\
\toprule
\multirow{3}{*}{TableBank-word }  & 10$\%$  & 80.5 & 92.5 & 87.7 & 87.1  \\
\cline{2-6}
& 30$\%$ & 88.3 & 95.7 & 93.1 & 92.1  \\
\cline{2-6}
& 50$\%$ & 91.5 & 96.7 & 95.2 & 94.5 \\
\midrule
 \multirow{3}{*}{TableBank-latex } & 10$\%$ & 63.7 & 93.5 & 71.6 & 74.3  \\
 \cline{2-6}
 & 30$\%$ & 82.8  & 96.4 & 93.4 & 89.0  \\
 \cline{2-6}
 & 50$\%$ & 85.3 & 96.2 & 94.4 & 91.4 \\
 \midrule
\multirow{3}{*}{TableBank-both }  & 10$\%$  & 84.2 & 95.8 & 93.1 & 90.1    \\
\cline{2-6}
 & 30$\%$ & 86.8 & 97.0 & 94.1 & 91.5 \\
 \cline{2-6}
 & 50$\%$ & 91.8 & 96.9 & 95.6 & 95.3  \\
\bottomrule
\end{tabular*}
\end{center}
\end{table} 
\begin{figure}[htp!]
\centering
\includegraphics[width=0.9\textwidth]{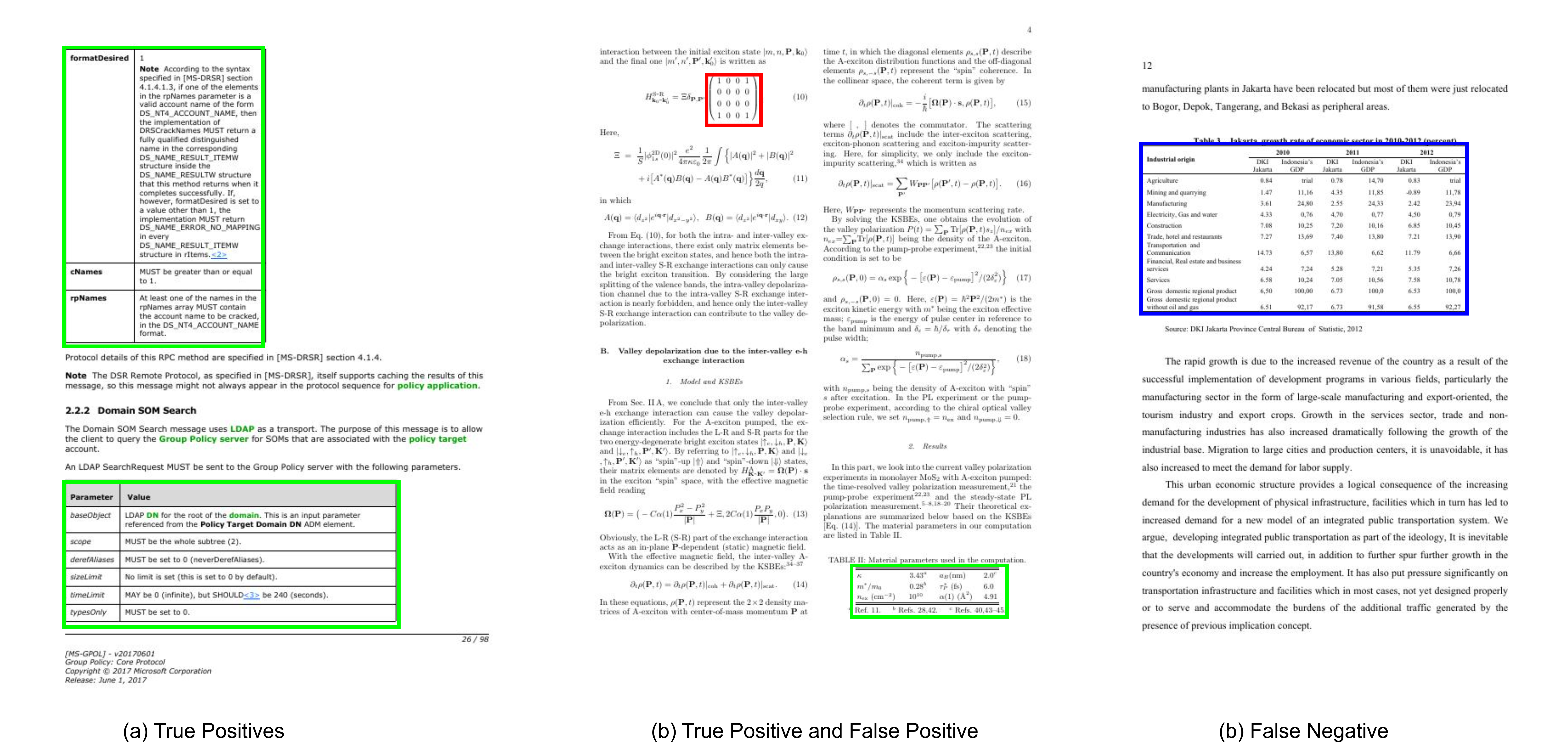}
\caption{Semi-supervised table detection results that employs deformable transformer on TableBank-both data split. Green color represents true positives, blue denotes false negatives and red shows false positives. Here, (a) indicates true positive detection results, (b) shows true positive and false positive detection results, and (c) gives false negative detection results.}\label{fig:tablebank-result}
\end{figure}

\begin{figure}
\centering
\includegraphics[width=1\textwidth]{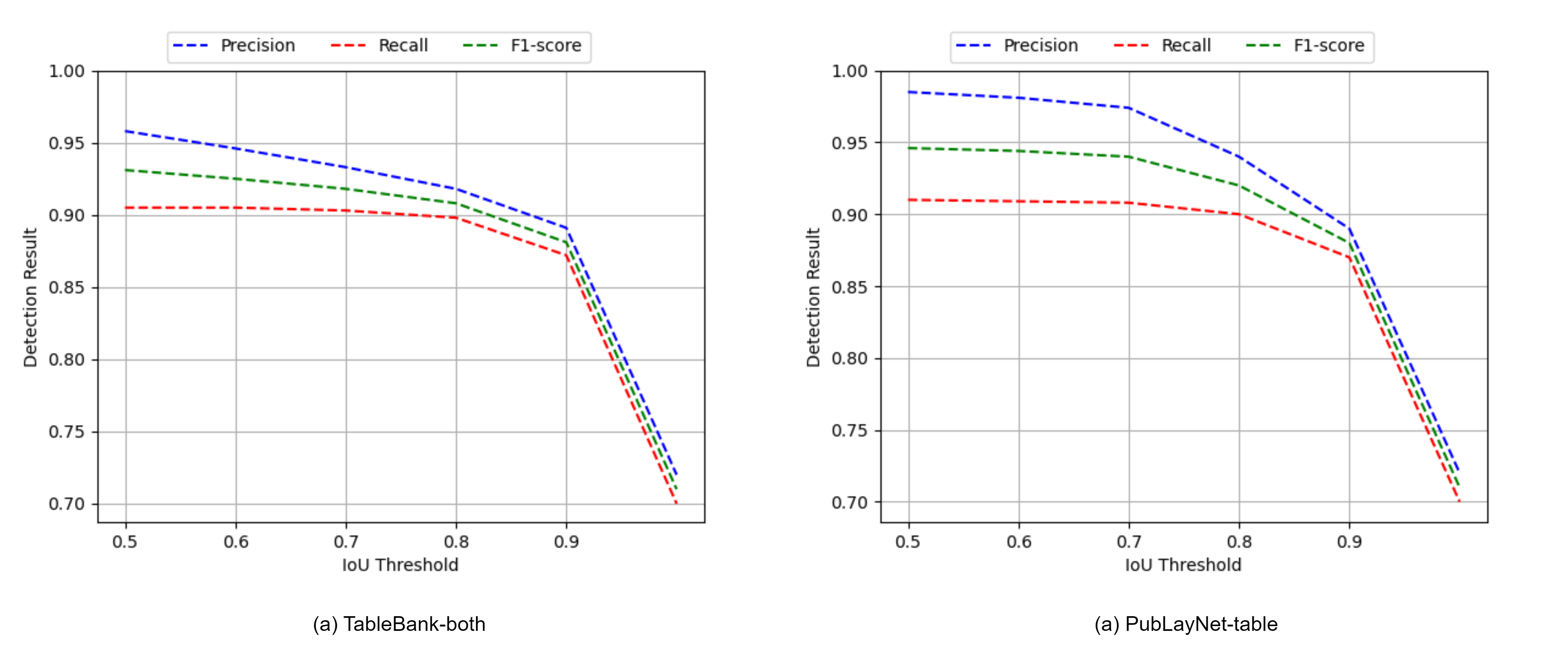}
\caption{A visual comparison of Precision, Recall and F1-Score of semi-supervised network that employs deformable transformer with ResNet-50 backbone on different IoU threshold values on 10\% labeled dataset of TableBank-both data split and PubLayNet table class dataset. Here, blue indicates precision results on different IoU threshold values, red shows recall results  on different IoU threshold values, and green represents F1-score results on different IoU threshold values.}\label{fig:iou}
\end{figure}

The qualitative analysis of semi-supervised learning for the TableBank-both data split is shown in Figure~\ref{fig:tablebank-result}. Part (b) of Figure~\ref{fig:tablebank-result} has a matrix with a similar structure as rows and columns, and the network detects the matrix as a table giving false positive detection results. Here, incorrect detection results indicate where the network fails to provide correct detection of table regions. Table~\ref{tab:tablebankIoU} gives the results of this semi-supervised approach on different IoU threshold values for all splits of the TableBank dataset on 10\% label data and the rest as unlabeled data.
A visual comparison of Precision, Recall and F1-Score of semi-supervised network that employs deformable transformer with ResNet-50 backbone on different IoU threshold values on 10\% labeled dataset of TableBank-both data split is shown in Figure~\ref{fig:iou}.
\begin{table}
\begin{center}
\caption{The performance comparison of semi-supervised network that employs deformable transformer with ResNet-50 backbone on different IoU threshold values on 10\% labeled dataset of TableBank-both data split.}\label{tab:tablebankIoU}
\renewcommand{\arraystretch}{1} 
\begin{tabular*}{\textwidth}
{@{\extracolsep{\fill}}ccccc@{\extracolsep{\fill}}}
\toprule
\textbf{Method} & 
\textbf{IoU} &
\textbf{Precision} & 
\textbf{Recall} &
\textbf{F1-score}     \\
\toprule
&  0.5 & 95.8  & 90.5 & 93.1\\
\cline{2-5}
    Semi-Supervised &  0.6 & 94.6 & 90.5 & 92.5\\
\cline{2-5}
Deformable-DETR+ResNet-50 & 0.7 & 93.3 & 90.3 & 91.8\\
\cline{2-5}
10\%  labels & 0.8 & 91.8 & 89.8 & 90.8\\
\cline{2-5}
& 0.9 & 89.1 & 87.2 & 88.1\\
\bottomrule
\end{tabular*}
\end{center}
\end{table}

\noindent\textbf{Comparisons with Previous supervised and semi-supervised approaches}
Table~\ref{tab:comptablebank} compares the deep learning-based supervised and semi-supervised networks on the ResNet-50 backbone. We also compare supervised deformable-DETR trained on 10\%, 30\% and 50\% TableBank-both data split label data with our semi-supervised approach that employs deformable transformer. It shows that our attention mechanism-based semi-supervised approach provides comparable results without using proposal generation process and post-processing steps such as Non-maximal suppression (NMS).
\begin{table}
\begin{center}
\caption{Performance comparison of previous supervised and semi-supervised approaches. Supervised Deformable-DETR and Faster R-CNN network trained on just 10\%, 30\% and 50\% data of TableBank-both dataset while semi-supervised networks used 10\%, 30\% and 50\% TableBank-both dataset as labeled and rest as unlabeled data using ResNet-50 backbone. Here, all results are represented on $mAP (0.5:0.95)$. The best threshold values are shown in bold.  }\label{tab:comptablebank}
\begin{tabular*}{0.9\textwidth}
{@{\extracolsep{\fill}}cccccc@{\extracolsep{\fill}}}
\toprule
\textbf{Method} & 
\textbf{Approach} &
\textbf{Detector} &
\textbf{$10\%$ } &
\textbf{$30\%$ }  & 
\textbf{$50\%$ }  \\
\toprule
 Ren et al. \cite{faster23}  & supervised  & Faster R-CNN & 80.1 & 80.6 & 83.3\\
 \midrule
Zhu et al. \cite{Deformable54}  & supervised  & Deformable DETR & 80.8 & 82.6 & 86.9\\
\midrule
STAC \cite{simplesemi76} &   semi-supervised & Faster R-CNN  & 82.4 & 83.8 & 87.1   \\
\midrule
Unbiased Teacher \cite{unbiasedT36} &   semi-supervised & Faster R-CNN  & 83.9 & 86.4 & 88.5\\
\midrule
Humble Teacher \cite{humbleTeacher6} &   semi-supervised & Faster R-CNN &  83.4 & 86.2 & 87.9 \\
\midrule
Soft Teacher \cite{softTeacher56} &   semi-supervised & Faster R-CNN  & 83.6 & \textbf{86.8} &  89.6 \\
\midrule
Our &   semi-supervised  & Deformable DETR  & \textbf{84.2} & \textbf{86.8} & \textbf{91.8} \\
\bottomrule
\end{tabular*}
\end{center}
\end{table} 

\subsection{PubLayNet}
In this subsection, we discuss the experimental results on PubLayNet table class dataset on different percentages of label data.  We also compare the transformer-based semi-supervised approach with previous deep learning-based supervised and semi-supervised approaches. Furthermore, we give results on 10\% PubLayNet dataset for all IoU threshold values. Table~\ref{tab:publaynet} provides the results of the semi-supervised approach that employs deformable transformer for PubLayNet table class on the different percentages of label data and rest as unlabeled data. Here, $AP_{50}$ value is 98.5\%, 98.8\%, and 98.8\% for 10\%, 30\% and 50\% label data, respectively.
\begin{table}
\begin{minipage}[b]{.49\textwidth}
\caption{Performance results for PubLayNet table class dataset. Here, mAP represents mean AP at the IoU threshold range of (50:95 ), $AP_{50}$ indicates AP at the IoU threshold of 0.5 and $AP_{75}$ denotes AP at the IoU threshold of 0.75. $AR_L$ indicates average recall for large objects.}\label{tab:publaynet}
\renewcommand{\arraystretch}{0.7} 
\begin{tabular*}{\textwidth}
{@{\extracolsep{\fill}}ccllll@{\extracolsep{\fill}}}
\toprule
\textbf{Dataset} & 
\textbf{Label-percent} &
\textbf{mAP} & 
\textbf{AP\textsuperscript{50}} &
\textbf{AP\textsuperscript{75}}  & 
\textbf{AR\textsubscript{L}}  \\
\toprule
\multirow{3}{*}{PubLayNet} & 10\% & 88.4 & 98.5 & 97.3 & 91.0  \\
\cmidrule{2-6}
& 30\% & 90.3 & 98.8  & 97.5 & 93.2  \\
\cmidrule{2-6}
 & 50\% & 92.8 & 98.8 & 97.3 & 96.0 \\
\bottomrule
\end{tabular*}
\end{minipage}\qquad
\begin{minipage}[b]{.48\textwidth}
\begin{center}
\caption{The performance comparison of semi-supervised network that employs deformable transformer with ResNet-50 backbone on different IoU threshold values on 10\% PubLayNet labeled Dataset.}\label{tab:publaynetIoU}
\renewcommand{\arraystretch}{0.7} 
\begin{tabular*}{\textwidth}
{@{\extracolsep{\fill}}lllll@{\extracolsep{\fill}}}
\toprule
\textbf{Method} & 
\textbf{IoU} &
\textbf{Precision} & 
\textbf{Recall} &
\textbf{F1-score}    \\
\toprule
 &  0.5 & 98.5 & 91.0 & 94.6\\
\cmidrule{2-5}
Semi-Supervised &  0.6 & 98.1 & 90.9 & 94.4  \\
\cmidrule{2-5}
Deformable-DETR & 0.7 & 97.4 & 90.8 & 94.0 \\
\cmidrule{2-5}
$10\%$ labels & 0.8 & 94.0 & 90.0 & 92.0\\
\cmidrule{2-5}
& 0.9 & 89.0 & 87.0 & 88.0\\
\bottomrule
\end{tabular*}
\end{center}
\end{minipage}
\end{table}

Furthermore, our semi-supervised network is trained on different IoU threshold values on 10\% of labeled PubLayNet Dataset. Table~\ref{tab:publaynetIoU} gives the results of the semi-supervised approach on different IoU threshold values for PubLayNet table class on 10\% label data and the rest as unlabeled data. A visual comparison of Precision, Recall and F1-score of the semi-supervised network that employs the deformable transformer network with ResNet-50 backbone on different IoU threshold values on 10\% labeled dataset of PubLayNet table class is shown in Figure~\ref{fig:iou}. Here, blue indicates precision results on different IoU threshold values on different IoU threshold values, red shows recall results, and green represents  F1-score results on different IoU threshold values.

\begin{table*}[htp!]
\begin{center}
\caption{ Performance comparison of previous supervised and semi-supervised approaches. Deformable-DETR and Faster R-CNN trained on just 10\%, 30\% and 50\% table data while semi-supervised networks used 10\%, 30\% and 50\% PubLayNet dataset as labeled and rest as unlabeled data. Here, all results are represented on $AP_{50}$ at the IoU threshold of 0.5. The best threshold values are shown in bold.} \label{tab:comppublaynet}
\renewcommand{\arraystretch}{1.1} 
\begin{tabular*}{\textwidth}
{@{\extracolsep{\fill}}cccccc@{\extracolsep{\fill}}}
\toprule
\textbf{Method} & 
\textbf{Approach} &
\textbf{Detector} & 
\textbf{$10\%$ } &
\textbf{$30\%$ }  & 
\textbf{$50\%$ }  \\
\toprule
 Ren et al. \cite{faster23} & supervised  & Faster R-CNN  & 93.6 & 95.6 & 95.9    \\
\midrule
Zhu et al. \cite{Deformable54} & supervised  & Deformable DETR  & 93.9 & 96.2 & 97.1    \\
\midrule

STAC \cite{simplesemi76} & semi-supervised & Faster R-CNN  & 95.8 & 96.9 & 97.8    \\
\midrule
Unbiased Teacher \cite{unbiasedT36} & semi-supervised & Faster R-CNN  & 96.1 & 97.4 & 98.1    \\
\midrule
Humble Teacher \cite{humbleTeacher6} & semi-supervised & Faster R-CNN  & 96.7 & 97.9 & 98.0  \\
\midrule
Soft Teacher \cite{softTeacher56} & semi-supervised & Faster R-CNN  & 96.5  & 98.1 & 98.5  \\
\midrule
Our & semi-supervised  & Deformable DETR & \textbf{98.5} & \textbf{98.8} & \textbf{98.8}    \\
\bottomrule
\end{tabular*}
\end{center}
\end{table*} 
\noindent\textbf{Comparisons with Previous supervised and semi-supervised approaches}
Table~\ref{tab:comppublaynet} compares the deep learning-based supervised and semi-supervised networks on PubLayNet table class using ResNet-50 backbone. We also compare supervised deformable-DETR trained on 10\%, 30\% and 50\% PubLayNet table class label data with our semi-supervised approach that employs the deformable transformer. It shows that our semi-supervised approach provides comparable results without using proposal and post-processing steps such as Non-maximal suppression (NMS).
\subsection{DocBank:}
In this subsection, we discuss the experimental results on DocBank dataset on different percentages of label data.  We compare the transformer-based semi-supervised approach with previous CNN-based semi-supervised approach in Table~\ref{tab:docbank}. 

\begin{table*}[htp!]
\begin{center}
\caption{ Performance comparison of previous semi-supervised approach and our Deformable-DETR based semi-supervised approach on DocBank dataset. Here, all results are represented on $mAP (0.5:0.95)$.} \label{tab:docbank}
\renewcommand{\arraystretch}{1.1} 
\begin{tabular*}{\textwidth}
{@{\extracolsep{\fill}}cccccc@{\extracolsep{\fill}}}
\toprule
\textbf{Method} & 
\textbf{Approach} &
\textbf{Detector} & 
\textbf{$10\%$ } &
\textbf{$30\%$ }  & 
\textbf{$50\%$ }  \\
\toprule
Soft Teacher \cite{softTeacher56} & semi-supervised & Faster R-CNN  &  72.3  & 74.4 & 81.5  \\
\midrule
Our & semi-supervised  & Deformable DETR & 82.5 & 84.9 & 87.1    \\
\bottomrule
\end{tabular*}
\end{center}
\end{table*} 
Furthermore, we also compare our semi-supervised approach on different percentages of label data with previous table detection and document analysis approaches for different datasets TableBank, PubLayNet, and DocBank in Table~\ref{tab:compall}. Although we cannot directly compare our semi-supervised approach with previous supervised document analysis approaches. However, we can observe that even with 50\% label data, we achieve comparable results with previous supervise approaches.
\begin{table*}[htp!]
\begin{center}
\caption{Performance comparison of previous supervised approaches for document analysis. Our semi-supervised network uses 10\%, 30\% and 50\% label data and rest as unlabeled data. Here, all results are represented on $mAP (0.5:0.95)$.} \label{tab:compall}
\renewcommand{\arraystretch}{1.1} 
\begin{tabular*}{\textwidth}
{@{\extracolsep{\fill}}ccccccc@{\extracolsep{\fill}}}
\toprule
\textbf{Method} & 
\textbf{Approach} & 
\textbf{ Labels} &
\textbf{TableBank} &
\textbf{PubLayNet }  & 
\textbf{DocBank} \\
\toprule
 CDeC-Net \cite{Agarwal52} & supervised & 100\%  &  96.5 & 97.8 & -    \\
\midrule
CasTabDetectoRS \cite{CasTab45} & supervised & 100\% & 95.3 & - & -     \\
\midrule
Faster R-CNN \cite{PubLayNet3} & supervised & 100\% & - & 90 & 86.3    \\
\midrule
VSR \cite{vsr45} & supervised & 100\% & - & 95.69 & 87.6   \\
\midrule
Our & semi-supervised & 10\%  & 84.2 & 88.4 & 82.5   \\
\midrule
Our & semi-supervised & 30\%  & 86.8 & 90.3 &  84.9  \\
\midrule
Our & semi-supervised & 50\%  & 91.8 & 92.8 &  87.1  \\
\bottomrule
\end{tabular*}
\end{center}
\end{table*} 
\subsection{ICDAR-19}
We also evaluate our method for table detection on the Modern Track A portion of the table detection dataset from the cTDaR competition at ICDAR 2019. We summarize the quantitative results of our approach at different percentages of label data and compare it with previously supervised table detection approaches in Table~\ref{tab:icdar19}. We evaluate results at higher IoU thresholds of 0.8 and 0.9. For a direct comparison with previous table detection approaches, we also evaluate our approach on 100\% label data. Our approach achieved a precision of 92.6\% and a recall of 91.3\% on the IoU threshold of 0.9 on 100\% label data.
\begin{table*}[htp!]
\begin{center}
\caption{Performance comparison between the proposed 
semi-supervised approach and previous state-of-the-art results on the dataset of ICDAR 19 Track A (Modern). } \label{tab:icdar19}
\renewcommand{\arraystretch}{1.1} 
\begin{tabular*}{\textwidth}
{@{\extracolsep{\fill}}cccccccc@{\extracolsep{\fill}}}
\toprule
\textbf{Method} &
\textbf{Approach} &
&
\textbf{IoU=0.8} &
&& 
\textbf{IoU=0.9} & \\
\cline{3-8} 
& &\textbf{Recall} &
\textbf{Precision} &
\textbf{F1-Score} &
\textbf{Recall} &
\textbf{Precision} &
\textbf{F1-Score} \\
\toprule
 TableRadar \cite{icdar19}  & supervised & 94.0 & 95.0 & 94.5 & 89.0 & 90.0 & 89.5    \\
\midrule
NLPR-PAL \cite{icdar19}  & supervised & 93.0 & 93.0 & 93.0 & 86.0 & 86.0 & 86.0    \\
\midrule
Lenovo Ocean \cite{icdar19}  & supervised & 86.0  & 88.0 & 87.0 & 81.0  & 82.0 & 81.5    \\
\midrule
CascadeTabNet \cite{Ayan29} & supervised & -  & - & 92.5 & - & - & 90.1    \\
\midrule
CDeC-Net \cite{Agarwal52} & supervised & 93.4  & 95.3 & 94.4 & 90.4 & 92.2 & 91.3    \\
\midrule
HybridTabNet \cite{Hyb65} & supervised & 93.3 & 92.0 & 92.8 & 90.5 & 89.5 & 90.2    \\
\midrule
Our & semi-supervised (50\%)  & 71.1 & 82.3 & 76.3 & 66.3 & 76.8 & 71.2  \\
\midrule
Our & supervised (100\%)  & 92.1 &  94.9 & 93.5 & 91.3  & 92.6 & 91.9  \\
\bottomrule
\end{tabular*}
\end{center}
\end{table*} 
\subsection{Ablation Study}
In this section, we validate the key design elements. Unless otherwise stated, all the ablation studies are conducted using a ResNet-50 backbone with 30\% labeled images from the PubLayNet dataset.

\noindent\textbf{Pseudo-Labeling confidence threshold}
In  Section~\ref{sec:semi-sup}, the threshold value (referred to as the confidence threshold) plays an important role in determining the balance between the accuracy and quantity of the generated pseudo-labels. As this threshold value increases, fewer examples will pass the filter, but they will be of higher quality. Conversely, a smaller threshold value will result in more examples passing but with a higher likelihood of false positives. The impact of various threshold values, ranging from 0.5 to 0.9, is presented in Table~\ref{tab:filter}. The optimal threshold value was determined to be 0.7 based on the results.

\begin{table}
\begin{minipage}[b]{.49\textwidth}
\caption{Performance comparison using different Pseudo-labeling confidence threshold values. The best threshold values are shown in bold.}\label{tab:filter}
\renewcommand{\arraystretch}{0.7} 
\begin{tabular*}{1\textwidth}
{@{\extracolsep{\fill}}cccc@{\extracolsep{\fill}}}
\toprule
\textbf{Threshold} & 
\textbf{AP} & 
\textbf{AP\textsuperscript{50}} &
\textbf{AP\textsuperscript{75}}    \\
\toprule
0.5   & 86.9 & 91.6 &  90.1   \\
\midrule
0.6 &  89.5 & 98.1 & 95.7  \\
\midrule
\textbf{0.7}  & \textbf{90.3} & \textbf{98.8} &  \textbf{97.5} \\
\midrule
0.8  &  89.4 & 97.2 & 95.3  \\
\midrule
0.9 & 87.9 & 96.3 &  94.5   \\
\bottomrule
\end{tabular*}
\end{minipage}\qquad
\begin{minipage}[b]{.48\textwidth}
\begin{center}
\caption{Performance comparison using different numbers of learnable queries to the decoder input. Here, best performance results are shown in bold.
}\label{tab:queries}
\renewcommand{\arraystretch}{0.7} 
\begin{tabular*}{1\textwidth}
{@{\extracolsep{\fill}}cccc@{\extracolsep{\fill}}}
\toprule
\textbf{N} & 
\textbf{AP} & 
\textbf{AP\textsuperscript{50}} &
\textbf{AP\textsuperscript{75}}     \\
\toprule
 3 & 61.4 & 69.7 & 62.6    \\
 \midrule
\textbf{30} & \textbf{90.3} & \textbf{98.8} & \textbf{97.5}   \\
\midrule
 50 & 89.4 & 90.3 & 85.4    \\
 \midrule
 100 & 88.4 & 89.7 & 83.9    \\
 \midrule
 300  & 78.5  & 94.7 & 90.2 \\
\bottomrule
\end{tabular*}
\end{center}
\end{minipage}
\end{table}

\begin{figure}
\centering
\includegraphics[width=0.9\textwidth]{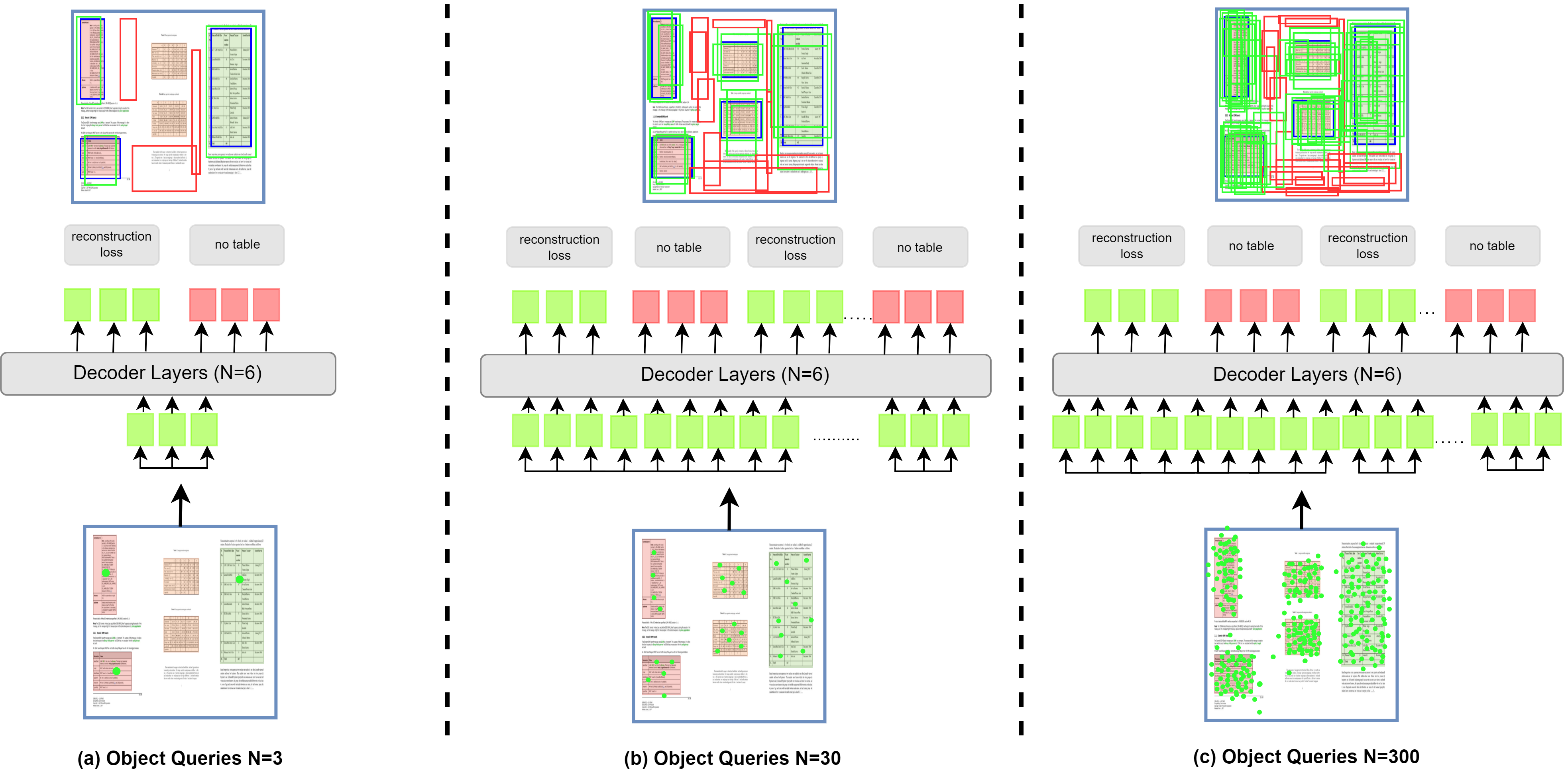}
\caption{Comparison of performance by variation of the number of object queries fed as input in the decoder of deformable DETR. Here, (a) takes N=3 object queries as input, (b) contains N =30 object queries as input, and (c) has N =300 object queries as input. The optimal performance is achieved by selecting the number of queries N to 30; deviating from this value results in a decrease in performance. Here, blue rectangles denote ground truth (GT), green rectangles indicate object class, and red rectangles show background class.}\label{fig:queries}
\end{figure}

\noindent\textbf{Influence of Learnable queries Quantity}
In our analysis, we investigate the impact of varying the number of queries fed as input in the decoder of deformable DETR. Figure~\ref{fig:queries} compares prediction results by varying the number of object queries fed as input in the decoder of deformable DETR. The optimal performance is attained when the number of queries N is set to 30; deviating from this value results in a decrease in performance. Table~\ref{tab:queries} presents and analyzes the result for varying object query quantities. Choosing a small value for N could result in the model failing to identify particular objects, negatively impacting its performance. On the other hand, selecting a large value for N may cause the model to perform poorly due to overfitting, as it would incorrectly classify certain regions as objects. Moreover, training complexity $O(N kc_i^2)$ of
this semi-supervised self-attention mechanism in the decoder of student-teacher module depends on the number of object queries and is subsequently improved as complexity is reduced by minimizing the number of object queries.
\section{Conclusion}
\label{sec:conclusion}
This paper introduces a semi-supervised approach that employs the deformable transformer for table detection in document images. The proposed method mitigates the need of large-scale annotated data and simplifies the process by integrating the pseudo-label generation framework into a streamlined mechanism. The simultaneous generation of pseudo-labels leads to a dynamic process known as the "flywheel effect", where one model continually improves the pseudo-boxes produced by the other model as the training progresses. The pseudo-class labels and pseudo-bounding boxes are improved in this framework using two distinct modules named student and teacher. These modules update each other by the EMA function to provide precise classification and bounding box predictions. The results indicate that this approach surpasses the performance of supervised models when applied to labeling ratios of 10\%, 30\%, and 50\% on TableBank all splits and the PubLayNet training data. Furthermore, when trained on the 10\% labeled data of PubLayNet, the model performed comparably to the current CNN-based semi-supervised baseline. In future, we aim to investigate the impact of the proportion of annotated data on the ultimate performance and develop models that function effectively with a minimal quantity of labeled data. Additionally, we intend to employ the transformer-based semi-supervised learning mechanism for table structure recognition task. 

%
%
%
\bibliographystyle{IEEEtran}
\bibliography{main}

\end{document}